%% file: neurips_2026.tex
\documentclass{article}

\usepackage[preprint]{neurips_2026}

\usepackage[utf8]{inputenc} 
\usepackage[T1]{fontenc}    
\usepackage{url}            
\usepackage{booktabs}       
\usepackage{amsfonts}       
\usepackage{nicefrac}       
\usepackage{microtype}      
\usepackage{xcolor}         

\usepackage{amsmath}
\usepackage{amssymb}
\usepackage{mathtools}
\usepackage{amsthm}
\usepackage{xspace}
\usepackage{enumitem}
\usepackage{multirow}
\usepackage{makecell}
\usepackage{graphicx}
\usepackage{subcaption}
\usepackage{caption}
\usepackage{wrapfig}
\usepackage[table]{xcolor}
\definecolor{lightblue}{rgb}{0.9, 0.95, 1}

\definecolor{scolor}{RGB}{111,168,220}
\definecolor{citecolor}{HTML}{229954}
\definecolor{eqc}{rgb}{1, 0, 0}
\usepackage[pagebackref=false,breaklinks=true,colorlinks=True,urlcolor=eqc,citecolor=citecolor,linkcolor=eqc,bookmarks=false,urlcolor=scolor]{hyperref}

\newcommand{\ourmethod}{World-VLA-Loop\xspace}
\newcommand{\ourdataset}{SANS\xspace}

\title{\ourmethod: Closed-Loop Learning of Video World Model and VLA Policy}

\author{Xiaokang Liu\thanks{Equal contribution} \quad Zechen Bai\footnotemark[1] \quad Hai Ci \quad Kevin Yuchen Ma \quad Mike Zheng Shou\footnotemark[2]\\
\\
Show Lab, National University of Singapore \\
\\
\url{https://showlab.github.io/World-VLA-Loop/}
}

\begin{document}
\footnotetext[2]{Corresponding authors} 

\maketitle

\begin{figure}[h]
\centering
\includegraphics[width=\columnwidth, trim=0 0 0 0 0, clip]{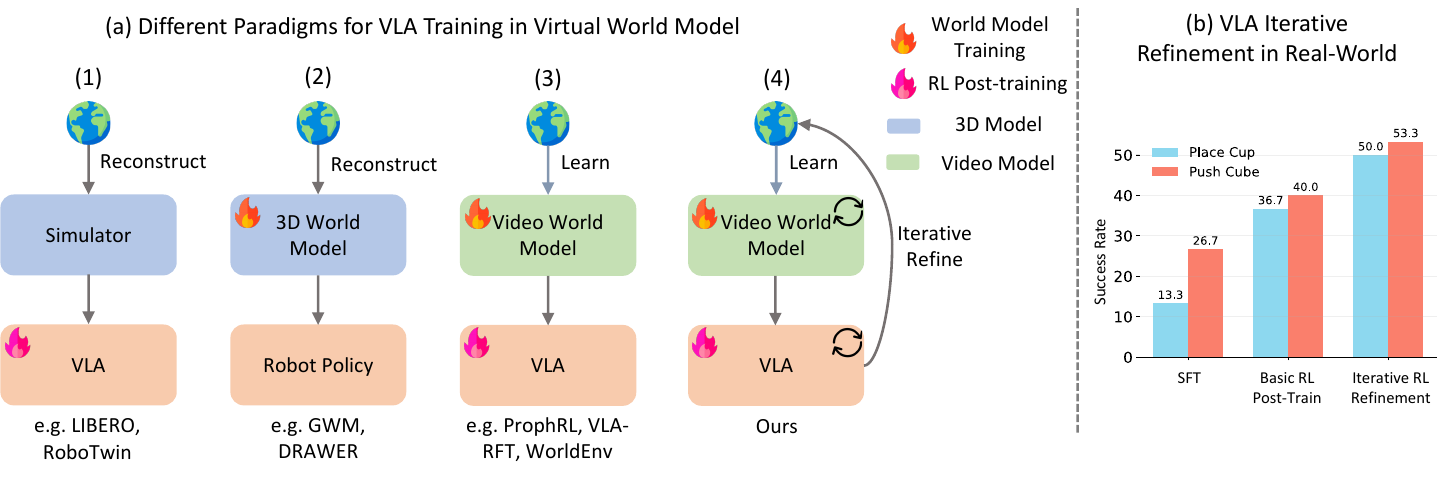} 
\caption{
(a) Paradigms for world-model-based VLA reinforcement learning. Comparison of existing methodologies: current approaches typically rely on reconstructing the environment within 3D world or training video world models that simulate the environment. To address the imprecise action-following inherent in existing video-based simulators, we propose \ourmethod, structured around two fundamental designs and a higher-level co-evolving paradigm. (b) We show that the real-world policy success rate is improved by 36.7\% and 26.6\% on two tasks after two iterations of joint optimization with VLA model and world model.
}
\label{fig:teaser}
\end{figure}

\input{sec/0_abstract_zechen}
\input{sec/1_intro_zechen}
\input{sec/2_related_work_zechen}

\input{sec/3_method}

\input{sec/4_experiments}
\input{sec/5_conclusion}

\bibliographystyle{plain}
\bibliography{reference}


\appendix

\input{sec/X_suppl}




\end{document}

%% file: sec/0_abstract_zechen.tex
\begin{abstract}

Reinforcement learning (RL) can refine Vision-Language-Action (VLA) policies beyond behavior cloning, but real-world RL remains expensive due to extensive rollouts, resets, supervision, and safety risks.
Action-conditioned video world models offer an option to train in virtual environments, yet they exhibit imprecise action following, particularly on subtle near-success failures.
Besides, they lack native reward signals for RL.
Computing rewards based on inaccurate visual predictions remain unreliable.
We introduce \ourmethod, structured around two foundational designs and a higher-level co-evolving paradigm.
We first curate \ourdataset, dedicatedly mixing successful and near-success trajectories to improve action--outcome alignment.
Then, we train a state-aware video world model that jointly predicts future frames and binary rewards from diffusion latents.
It couples reward estimation to the generator rather than a separate module, and in turn, benefits visual prediction.
Since VLA behavior shifts during RL, a fixed simulator can misalign with the updated policy, \ourmethod therefore closes the loop by using the refined world model for iterative VLA post-training while feeding rollouts from each improved policy back to augment and fine-tune the world model.
Across simulation and real-robot experiments, \ourmethod substantially improves VLA performance while reducing reliance on costly physical interaction.

\end{abstract}

%% file: sec/1_intro_zechen.tex
\section{Introduction}
\label{sec:intro}

Vision-Language-Action (VLA) models have emerged as a dominant paradigm for robotic manipulation by leveraging large language model priors to map natural language directly to low-level control \cite{black2024pi_0, kim2024openvla, kim2025fine, liu2025diffusion}.
Beyond behavior cloning, which requires large-scale human demonstration collection, recent works \cite{lu2025vla, li2025simplevla, intelligence2025pi} integrate reinforcement learning (RL) to refine policies through interaction.
Despite this promise, existing RL post-training remains largely confined to simulated environments \cite{liu2023libero, tao2024maniskill3}, because real-world RL incurs substantial costs: extensive physical rollouts, safety risks during exploration, and operational overhead for resets and supervision.

A promising way to mitigate these challenges is to treat learned world models as virtual environments.
As shown in Fig.~\ref{fig:teaser}, we group such models into three paradigms:
(1) \emph{Handcrafted digital twins} \cite{liu2023libero, chen2025robotwin}, which rely on manual asset creation and physics engine modeling but often lack the photorealism and physical fidelity needed for real-world adaptation;
(2) \emph{3D-based reconstruction} \cite{lu2025gwm, xia2025drawer}, which use geometric 3D representations but struggle to generalize across diverse environments, and rarely support rich stochastic exploration;
(3) \emph{Action-conditioned video world models} \cite{zhang2025reinforcing, li2025vla, xiao2025world}, which leverage pretrained priors from large-scale video and offer high photorealism and stronger scene generalization.

\begin{figure}[t]
  \centering
   \includegraphics[width=0.9\linewidth]{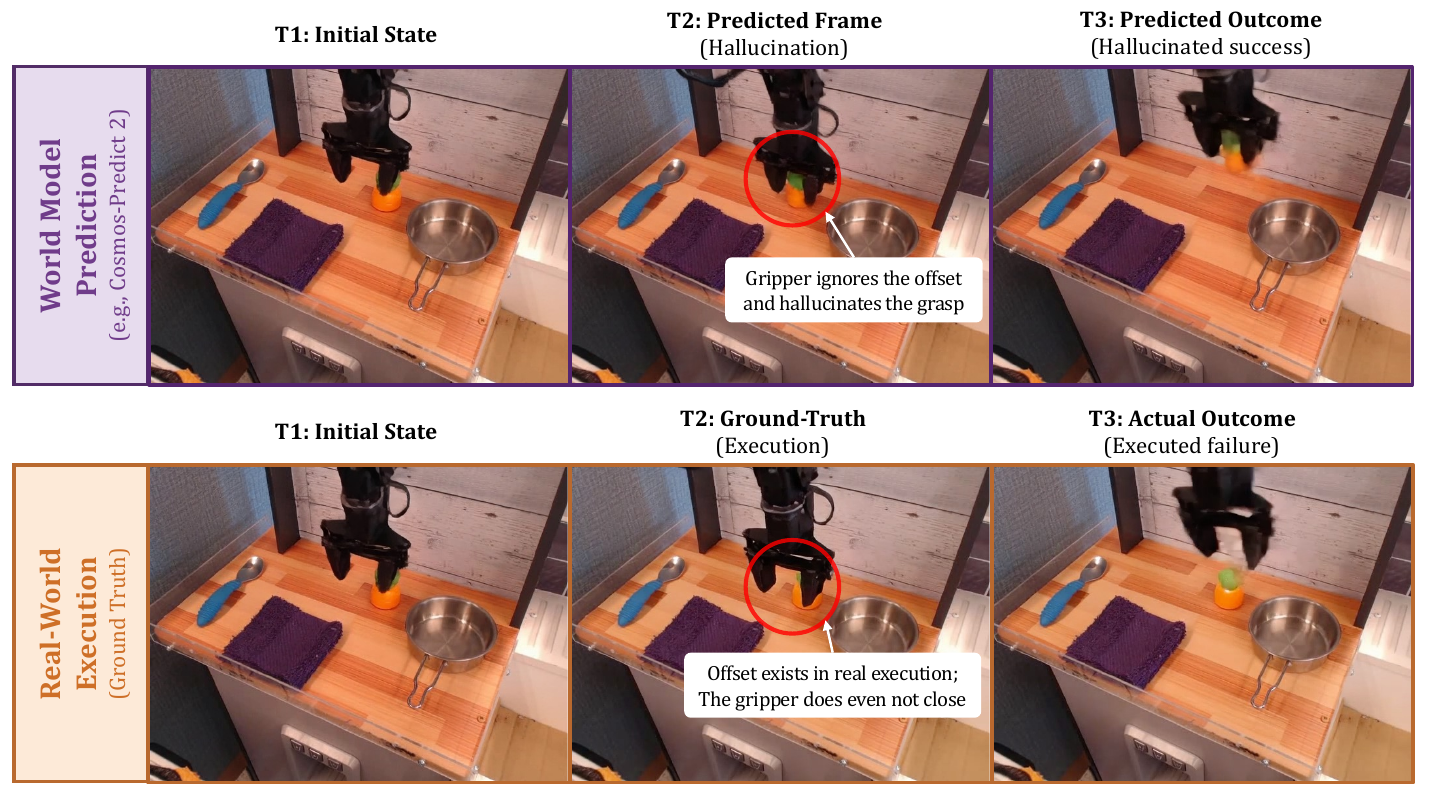}
   \caption{Current world models struggle to accurately simulate failure cases stemming from minor action errors, largely because they inadequately model fine-grained interaction dynamics and precise action conditioning. Transparent overlays denote ground-truth gripper trajectories, illustrating cases where the robot fails to grasp the object.}
   \label{fig:cosmos-failure-case}
   \vspace{-8pt}
\end{figure}

Video world models are attractive, yet they face two critical limitations.
First, our empirical analysis shows that they often follow actions imprecisely, so predicted trajectories diverge from actual execution outcomes.
This is particularly prominent in \emph{near-success} failure cases, where the robot fails to achieve a specific goal due to minor action errors.
As shown in Fig.~\ref{fig:cosmos-failure-case}, models such as Cosmos-Predict 2 \cite{ali2025world} frequently hallucinate successful outcomes even under erroneous actions, reflecting weak grounding in fine-grained physical dynamics.
Second, they do not natively provide reward signals for RL.
Although recent work~\cite{xiao2025world} constructs rewards from visual predictions, such rewards remain unreliable when the underlying video predictions are inaccurate, especially in the (near-)success cases.

We address these limitations with two complementary \emph{foundational} designs.
Specifically, to improve action--outcome alignment, we curate the \textbf{\ourdataset} dataset, which combines \textbf{S}uccessful \textbf{A}nd \textbf{N}ear-\textbf{S}uccess trajectories so the model learns sharper boundaries between success and subtle failure.
To obtain reliable learning signals for RL, we introduce a \textbf{state-aware video world model} that jointly predicts future frames and binary reward signals (success versus failure).
At training time, joint supervision on video and rewards further sharpens action following by encouraging a clear success--failure boundary in latent space.
At inference time, the predicted rewards serve naturally as RL rewards.
Experimental results show that these designs substantially improve the video world model and, in turn, enable effective RL post-training of VLA policies.

Despite the promising results achieved by the above designs, a separate challenge appears at the level of the \emph{full} learning system.
With RL post-training, \emph{VLA behaviors shift} as the policy improves: updated policies induce different mistakes, and therefore stress the world model in ways that differ from the data on which the world model was initially fit.
A simulator that remains static can become misaligned with the post-update policy's failure modes, so both predicted trajectories and rewards may drift.
We elevate this observation into \ourmethod, a higher-level \emph{co-evolving} paradigm built on top of the two foundations.
\ourmethod closes the loop between simulator and policy: the refined world model supports iterative VLA post-training inside the virtual environment, while rollouts from each updated policy are fed back to augment training and fine-tune the world model. 
Alongside the foundational dataset and world-model design, this closed-loop co-evolution between world modeling and policy learning turns reliable one-shot simulation into an iterative system that stays aligned with the policy under training.
Our contributions are as follows:
\begin{itemize}[nosep]
    \item We characterize \emph{near-success} failure modes in video world models and curate \ourdataset to improve action--outcome alignment.
    \item We develop a state-aware world model that achieves strong action-following precision and native reward prediction by joint reward-and-video supervision on diffusion latents.
    \item We introduce \ourmethod, a closed-loop framework that establishes a co-evolving cycle between world model and VLA policy learning through iterative refinement.
    \item We conduct evaluation across both simulation and real-world settings, showing that \ourmethod significantly improves VLA performance across simulation and physical deployment while reducing reliance on costly real-world interaction.
\end{itemize}

%% file: sec/2_related_work_zechen.tex
\section{Related Work}
\label{sec:related_work}

\paragraph{VLA for Robot Manipulation}
By leveraging the strong priors of foundation VLMs, VLA has emerged as a mainstream paradigm for robot manipulation.
Pretrained VLMs offer strong generalization and are typically fine-tuned on robotic demonstration data to adapt to action spaces \cite{driess2023palm, karamcheti2024prismatic, kim2024openvla, kim2025fine, black2024pi_0}.
However, the dominant training paradigm remains imitation learning, which suffers from data scarcity and compounding errors at test time, ultimately limiting generalization.
To mitigate these limitations, reinforcement learning has become an attractive option, enabling policies to learn from online rollouts beyond static human demonstrations.
Representative works include VLA-RL~\cite{lu2025vla}, RL-VLA~\cite{liu2025can}, SimpleVLA-RL~\cite{li2025simplevla}, RLinf-VLA~\cite{zang2025rlinf}, among others.
A fundamental bottleneck for adapting RL post-training to the real world is prohibitive cost: extensive rollouts, operational overhead for resets and supervision, and safety risks during exploration.
To address this, our framework eliminates the costs of physical interaction by simulating the real-world environment within a high-fidelity video world model, enabling efficient policy RL.

\paragraph{Video World Models for Robotics}
With recent advances in video generation~\cite{wan2025wan, blattmann2023stable}, controllable video-based world models have become increasingly feasible.
Unlike standard generative models that rely solely on text prompts, world models incorporate action sequences as conditioning signals.
Many works leverage the spatiotemporal priors of large pretrained video models to enable action-conditioned simulation in domains such as gaming~\cite{bruce2024genie, yu2025gamefactory}, autonomous driving~\cite{ni2025maskgwm, hu2024drivingworld}, and robotics~\cite{li2025vla, xiao2025world, quevedo2025evaluating, ali2025world, zhang2025reinforcing, zhu2025wmpo, guo2026vlaw, jiang2026wovr}.
Despite this progress, empirical evidence suggests that current video world models still struggle with precise action following, yielding misalignment between predicted frames and the outcomes implied by the actions.
To integrate world models into VLA training, particularly RL post-training, practitioners need not only a visual environment but also a reward signal.
Recent work explores several designs.
VLA-RFT~\cite{li2025vla} employs a visual-similarity-based heuristic reward.
World-Env~\cite{xiao2025world} uses a VLM to estimate success from the world model's visual predictions.
WMPO~\cite{zhu2025wmpo} implements the reward model with a video encoder and a linear head.
Most such approaches rely on an additional trained module and on visual predictions that may themselves be unreliable.
We instead predict rewards directly from the world model's latents.
Joint training not only enables end-to-end reward prediction but also benefits the visual prediction branch.
Concurrent works, including WoVR~\cite{jiang2026wovr} and VLAW~\cite{guo2026vlaw}, also identify the imagination-outcome misalignment issue of world models and propose the similar idea of co-evolving of world model and VLA.
However, they still rely on external reward model yet does not investigate the intrinsic reward and its relation with the world state.

%% file: sec/3_method.tex
\begin{figure*}[t]
  \centering
   \includegraphics[width=\linewidth]{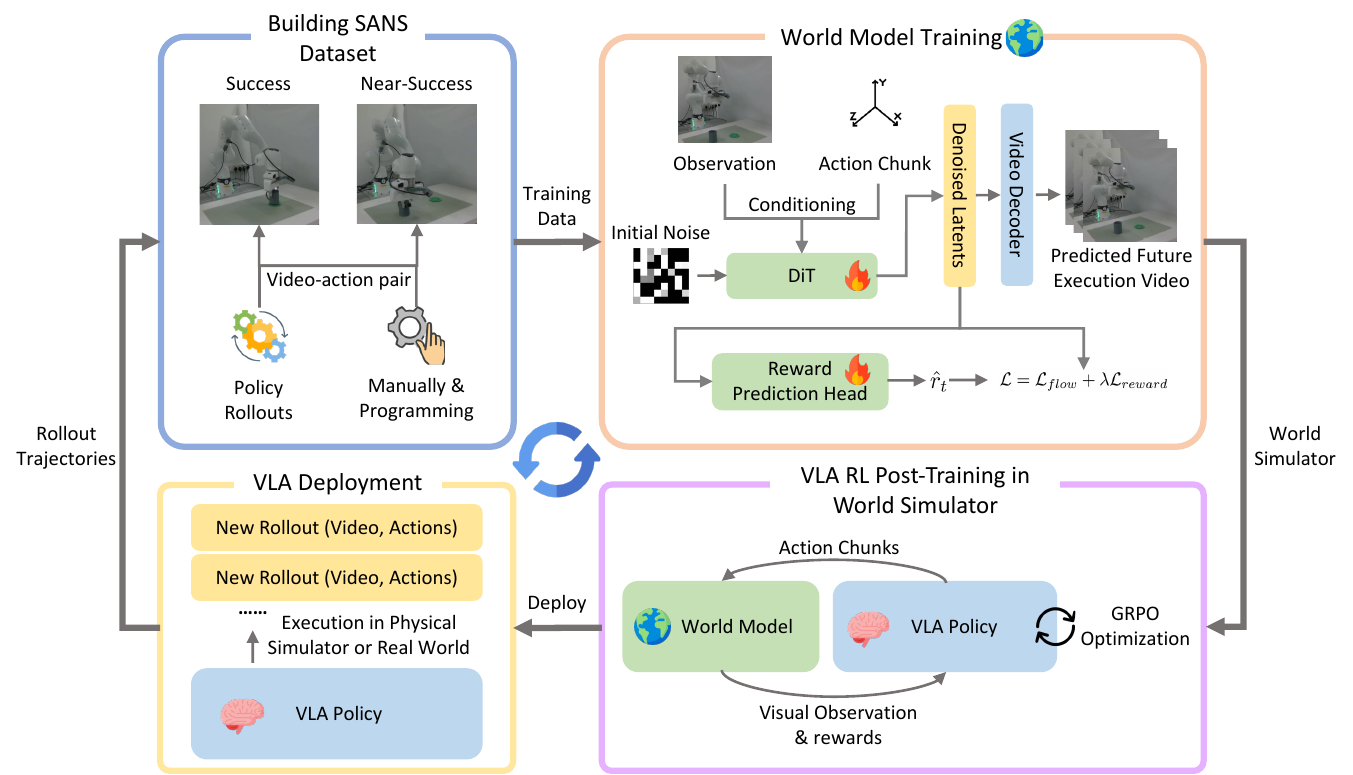}
   \caption{Full pipeline of our proposed framework. The process comprises four phases: (1) Curating the \ourdataset dataset via manual teleoperation and policy rollouts; (2) Pretraining the action-conditioned world model on \ourdataset with joint reward and video supervision; (3) Executing VLA policy rollouts within the world model to perform GRPO optimization; and (4) Deploying the refined policy to collect new failure and success data for further \ourdataset augmentation. This cycle enables the joint optimization of the world model and the VLA policy, iteratively enhancing both performance.}
   \label{fig:pipeline}
\end{figure*}

\section{Method}

\label{sec:method}

As shown in Figure~\ref{fig:pipeline}, our framework comprises four phases. We first curate a dataset of success and near-success trajectories across diverse outcomes. We then train an action-conditioned video world model with reward supervision on top of Cosmos-Predict \cite{ali2025world}. The trained model can be efficiently adapted to unseen scenarios using a few demonstrations, and serves as a virtual environment for policy evaluation and RL post-training. Real-world policy rollouts are further used to augment the dataset. We detail each stage below.

\subsection{Success and Near-Success Dataset}

Existing open robotic datasets with action-trajectory annotations, regardless of their diversity, predominantly focus on successful executions as they are primarily curated for imitation learning. This convention restricts the diversity required to train robust world models, which consequently struggle to simulate physically plausible outcomes across diverse failure modes. While recent works like RoboFAC \cite{lu2025robofac} and AHA \cite{duan2024aha} have begun exploring failure data, they are designed for QA-style reasoning and lack action annotations. Moreover, these datasets are largely confined to simulation, leaving its practical utility underexplored.

In this paper, we introduce the Success and Near-Success Dataset (\ourdataset), which leverages trajectories that nearly accomplish the target goal or sub-goal but fail due to minor inaccuracies in end-effector positioning. Such data is critical for training robust world models for two reasons: (1) these trajectories are difficult to distinguish from successful executions, they force the world model to focus on fine-grained nuances in spatial dynamics; and (2) since robot policies frequently exhibit these "near-success" behaviors, including them ensures the virtual environment more accurately reflects the actual failure modes encountered during policy rollouts. We curate the \ourdataset dataset across ManiSkill \cite{tao2024maniskill3}, LIBERO \cite{liu2023libero}, and our own real-world robotic settings.



For ManiSkill \cite{tao2024maniskill3}, we collect success trajectories using simple control policies with ground-truth object poses, and generate failures via pose perturbations. We also capture natural failure modes from policy rollouts, a strategy we follow in LIBERO using OpenVLA-OFT \cite{kim2025fine}, and in real-world settings via both teleoperation and OpenVLA-OFT rollouts. Each trajectory includes video, actions, and step-wise binary reward signals indicating success.

Using this pipeline, we build a large ManiSkill dataset for world model pretraining, comprising ~35k video-action pairs across 23 tasks with diverse success and failure outcomes.
For LIBERO and real-world scenarios, we collect smaller \ourdataset datasets (roughly 50 success and 50 failure trajectories per task), and show that the pretrained model can be efficiently transferred to these unseen tasks.

\subsection{State-aware Video World Simulator}

We build our video world simulator on top of Cosmos-Predict 2 \cite{ali2025world}. Cosmos-Predict 2 is pretrained on a large corpus of embodiment-related video data and can be further adapted to incorporate robot action conditions. Given the first $h$ observed frames $x_0, \dots, x_{h-1}$ and next $T$ timesteps of robot actions $a_1, \dots, a_T \in \mathbb{R}^6 \cup \{0, 1\}$ (represented as 6-DoF end-effector poses together with the gripper open/close state) the model synthesizes future execution frames for the next $T$ steps, denoted as $x_{h}, \dots, x_{h+T-1}$.

Cosmos-Predict utilizes a Diffusion Transformer (DiT) backbone to predict future video chunks autoregressively. To accommodate the action modality, an action embedder MLP maps each input pose into a latent tensor. These action embeddings are integrated into the DiT modules by adding them directly to the diffusion timestamp embeddings. This design enables the model to effectively ground its visual predictions in the provided control sequence while leveraging its strong pretrained generative priors.

To enable reward prediction without relying on external reward models, we augment the original generative model with a reward prediction head, as illustrated in Figure~\ref{fig:pipeline}. After the DiT model produces diffusion latents, we add a projection layer that maps these latents to a sequence of reward values, supervised by ground-truth rewards collected in the \ourdataset dataset. Specifically, for each sampled timestep $t$, let $z_t$ denote the denoised latent at the final step. The reward head $\phi$ predicts the scalar reward $\hat{r}_t = \phi(z_t)$, where $\phi$ is implemented as a lightweight MLP. During training, we jointly optimize the reward prediction head alongside the original flow matching loss \cite{lipman2022flow}:
$$
\mathcal{L} = \mathcal{L}_{flow} + \lambda \sum_{t=1}^{T} \| \hat{r}_t - r_t \|^2,
$$
where the weighting factor $\lambda$ is modulated according to the sampled noise level following the EDM framework \cite{karras2022elucidating}. This ensures the reward head remains robust to the high-variance latents encountered during the early stages of the denoising process.

This design brings two key advantages: \textbf{(1)} it provides a reliable reward generation mechanism within the world model. Since rewards are predicted based on the generated video latent states, they are intrinsically aligned with the actual visual outcomes and tend to be more accurate than low-level heuristic proxies or external VLM-based reward models. \textbf{(2)}, joint training of the reward head and the video model encourages the generator to better distinguish successful versus failed execution outcomes under different action conditions. This leads to more accurate future video predictions—a crucial capability often lacking in existing video world models.

In practice, we initialize from the Cosmos-Predict checkpoint, first pretrain it on the collected ManiSkill \ourdataset dataset, and then fine-tuning on a small set of success and near-success demonstrations for new scenarios. The ManiSkill-pretrained model provides strong action following ability, while fine-tuning imparts knowledge of new environments, object textures, and task-specific interactions.

\subsection{World Simulator for Policy RL Post-Training}
\label{method:world-simulator-deployment}

Once trained to precisely generate execution outcomes conditioned on action inputs, our video world model serves as a reliable virtual environment for downstream policy evaluation and RL post-training.



We adopt OpenVLA-OFT \cite{kim2025fine} as the base policy and build a GRPO-based post-training pipeline following SimpleVLA-RL \cite{li2025simplevla}. For each target task with a sparse \ourdataset dataset (typically fewer than 50 success and 50 near-success trajectories), we first fine-tune both the world model and policy to the specific scenario. During RL, the learned world model replaces physics-based simulators, autoregressively generating multi-step visual observations and reward predictions conditioned on policy actions. These predictions are fed back to the policy in a closed loop to enable full rollout generation. The world model also provides step-wise rewards, which are thresholded into binary success signals and used to compute advantages over groups of rollouts for policy updates.

As depicted in Figure~\ref{fig:pipeline}, a key advantage of our method is the establishment of a comprehensive, iterative joint-optimization framework for both the world model and the VLA policy. Following each RL update, we collect success and near-success rollouts generated by the newly trained VLA policy when deployed in its target environment. These rollouts are then used to augment the \ourdataset dataset for subsequent iterations of world model fine-tuning. This iterative refinement ensures that the world model progressively improves its ability to synthesize precise execution outcomes, which in turn enhances the effectiveness of the next RL phase for the VLA policy. In the experimental section, we validate that this joint-optimization process does yield superior VLA RL performance.

%% file: sec/4_experiments.tex
\section{Experiments}

\label{sec:experiments}

The following section evaluates our framework across multiple dimensions, validating the effectiveness and broader applicability of the proposed approach. Our results demonstrate that:
\textbf{(1)} The world model achieves an average visual alignment of 88.5\% and a reward alignment of 87.25\%, establishing it as a high-fidelity simulator for VLA policy rollouts.
\textbf{(2)} Through RL post-training within our simulator, the OpenVLA-OFT policy achieves an average success rate improvement of 12.7\% on the LIBERO benchmarks, as well as 23.4\% and 13.3\% in real-world scenarios.
\textbf{(3)} The proposed iterative refinement process further increases real-world task accuracy by an average of 13.3\% relative to the first RL checkpoint, underscoring the efficacy of the joint optimization cycle.

\subsection{Experiment Settings}

\textbf{Experiment Scenarios.} We conduct experiments on the LIBERO benchmark for simulated tasks, complemented by a self-constructed laboratory setup to evaluate real-world applicability. In both settings, the configuration consists of a Franka research arm \cite{franka_emika} and a single third-person-view RealSense D435 camera \cite{keselman2017intel} mounted on a fixed position to provide observation input. For the real-world experiments, we design two major tasks \textit{Pick and Place Cup} and \textit{Pushing Cube}. We use a unified chunk size of 24 for all tasks in simulation and real-world. For more implementation details and real-world settings, please refer to Appendix~\ref{sec:implementation-details}.

\textbf{Evaluation Methods.} We evaluate our pipeline from two perspectives. First, we assess the generative performance of the \ourmethod world model, focusing on visual fidelity and action-following precision. Building upon this, we quantify the model’s utility as a simulator by measuring the success rate improvements of a VLA policy during RL post-training.

\subsection{World Model Evaluation}

To ensure that the world model provides reliable generation suitable for accurate policy rollouts, we first evaluate its generative quality. Specifically, given a test dataset of image-actions pairs, the world model takes an initial image and a sequence of absolute robot end-effector poses as input. It then autoregressively predicts the video frames corresponding to the execution of those actions. We evaluate the resulting generations based on two primary dimensions: video quality and action-following accuracy.

\textbf{Video Quality}
Since the generated frames serve as the major visual observations for the policy, in order to maintain the stable behavior of VLA, ensuring high visual fidelity is crucial. As indicated in Table~\ref{tab:video-quality}, \ourmethod world model achieves high-quality generation results across both simulation and real-world scenarios, yielding high-fidelity results with negligible structural and perceptual distortion. The consistency between the LIBERO and real-world metrics demonstrates that our world model generalizes effectively across different environments, providing a reliable visual foundation for downstream policy rollouts.

\begin{table*}[t]
\centering
\caption{Video generation performance. $\uparrow$ indicates higher is better; $\downarrow$ indicates lower is better.}
\resizebox{0.55\linewidth}{!}{
\begin{tabular}{lcccc}
\toprule
\textbf{Scenario} & \textbf{SSIM} $\uparrow$ & \textbf{PSNR} $\uparrow$ & \textbf{LPIPS} $\downarrow$ & \textbf{MSE} $\downarrow$\\
\midrule
LIBERO & 0.90 & 26.57 & 0.031 & 0.0024 \\
Real-World & 0.91 & 29.61 & 0.059 & 0.0019  \\
Average & 0.91 & 28.09 & 0.045 & 0.0022  \\
\bottomrule
\end{tabular}
}
\label{tab:video-quality}
\end{table*}

\begin{table*}[t]
\centering
\caption{Outcome Alignment Performance across specific LIBERO and Real-World tasks. For each task we evaluate 50 samples within   set, and report the percentage of samples where the predicted success/failure matches the ground truth.}
\label{tab:generation-accuracy}
\resizebox{\columnwidth}{!}{%
\begin{tabular}{@{}lccccccccc@{}}
\toprule
\multirow{2}{*}{\textbf{Metric}} & \multicolumn{2}{c}{\textbf{LIBERO-Object}} & \multicolumn{2}{c}{\textbf{LIBERO-Goal}} & \multicolumn{2}{c}{\textbf{LIBERO-Spatial}} & \multicolumn{2}{c}{\textbf{Real-World}} \\ \cmidrule(lr){2-3} \cmidrule(lr){4-5} \cmidrule(lr){6-7} \cmidrule(lr){8-9}
 & \textbf{Task 1} & \textbf{Task 2} & \textbf{Task 1} & \textbf{Task 2} & \textbf{Task 1} & \textbf{Task 2} & \textbf{Place Cup} & \textbf{Push Cube} \\ \midrule
Visual Alignment & 92\% & 90\% & 94\% & 78\% & 86\% & 94\% & 90\% & 84\% \\
Reward Alignment & 88\% & 90\% & 90\% & 76\% & 88\% & 94\% & 94\% & 78\% \\ \bottomrule
\end{tabular}
}
\vspace{-4mm}
\end{table*}

\begin{table*}[t]
\centering
\caption{Success rate of OpenVLA-OFT before and after RL training. Success rates are computed across 500 rollouts for the LIBERO suites and 30 physical rollouts for our real-world experiments. The Oracle results are obtained by RL training inside LIBERO simulator with ground truth videos and rewards, and the numbers are averaged across the entire tasksuite.}
\label{tab:vla-rl-improvements}
\setlength{\aboverulesep}{0pt}
\setlength{\belowrulesep}{0pt}
\renewcommand{\arraystretch}{1.3} 
\resizebox{\columnwidth}{!}{%
\begin{tabular}{@{\hspace{6pt}}lccccccccc@{\hspace{6pt}}} 
\toprule
\multirow{2}{*}{\textbf{Model}} & \multicolumn{2}{c}{\textbf{LIBERO-Object}} & \multicolumn{2}{c}{\textbf{LIBERO-Goal}} & \multicolumn{2}{c}{\textbf{LIBERO-Spatial}} & \multicolumn{2}{c}{\textbf{Real-World}} \\ \cmidrule(lr){2-3} \cmidrule(lr){4-5} \cmidrule(lr){6-7} \cmidrule(lr){8-9}
 & \textbf{Task 1} & \textbf{Task 2} & \textbf{Task 1} & \textbf{Task 2} & \textbf{Task 1} & \textbf{Task 2} & \textbf{Place Cup} & \textbf{Push Cube} \\ \midrule
SFT Base & 73.9\% & 73.9\% & 91.9\% & 86.1\% & 83.9\% & 87.9\% & 13.3\% & 26.7\% \\
RL Post-Training (Ours) & 97.9\% & 91.9\% & 100\% & 96.2\% & 93.9\% & 94.0\% & 36.7\% & 40.0\% \\
\rowcolor{lightblue} $\Delta$ vs SFT & +24.0\% & +18.0\% & +8.1\% & +10.1\% & +10.0\% & +6.1\% & +23.4\% & +13.3\%\\
\midrule
RL Post-Training (Oracle) & 98.7\% & 98.7\% & 98.8\% & 98.8\% & 98.2\% & 98.2\% & - & - \\
\bottomrule
\end{tabular}
}
\end{table*}

\textbf{Generation Accuracy}
A critical requirement for a world simulator is the ability to faithfully reflect the causal consequences of an action trajectory. While pixel-level metrics like MSE provide a baseline, a more robust evaluation measures the alignment between generated and ground-truth outcomes. To this end, we categorize the final state of each trajectory as either a success or failure and calculate the alignment rate—the percentage of samples where the predicted outcome matches the ground truth. Since our world model simultaneously generates video frames and scalar rewards, we report two distinct alignment metrics:
\begin{itemize}[nosep]
    \item \textit{Visual Alignment}: Measures the faithfulness of the pixel-level generation. Success or failure is determined by evaluating the final generated video frames to judge the task outcome.
    \item \textit{Reward Alignment}: Measures the world model reward prediction. A predicted trajectory is classified as a success if the generated reward exceeds a threshold of $0.9$.
\end{itemize}

As shown in Table~\ref{tab:generation-accuracy}, our method effectively distinguishes between successful and failed trajectories across both simulation and real-world scenarios, achieving an average alignment accuracy exceeding 80\% for both metrics.
Furthermore, the outcomes classified via reward alignment are highly consistent with those from visual alignment.
This strong correlation underscores the reliability of the internal reward prediction head and demonstrates its learned success criteria are well-aligned with human judgment.

\subsection{World Simulator for VLA Post-training}

\textbf{Basic RL Training} Leveraging its reliable generative performance, \ourmethod world model serves as a learned simulator for VLA policy post-training. For each task, we first conduct task-specific finetuning for both the OpenVLA policy and the world model. Subsequently, the policy is deployed within the world model environment for RL post-training. We adopt the RL pipeline from SimpleVLA-RL \cite{li2025simplevla}, adapting the rollout process to operate entirely within our neural simulator. In this configuration, only the initial frame is sourced from the original dataset; all subsequent observations are autoregressively generated by the world model conditioned by the policy's predicted actions. Finally, the RL reward is derived from the world model’s reward head, utilizing a thresholding mechanism to determine the binary success signals required for group relative optimization \cite{shao2024deepseekmath}.

\begin{figure*}[t]
    \centering
    \begin{subfigure}[b]{0.24\textwidth}
        \centering
        \includegraphics[width=\linewidth]{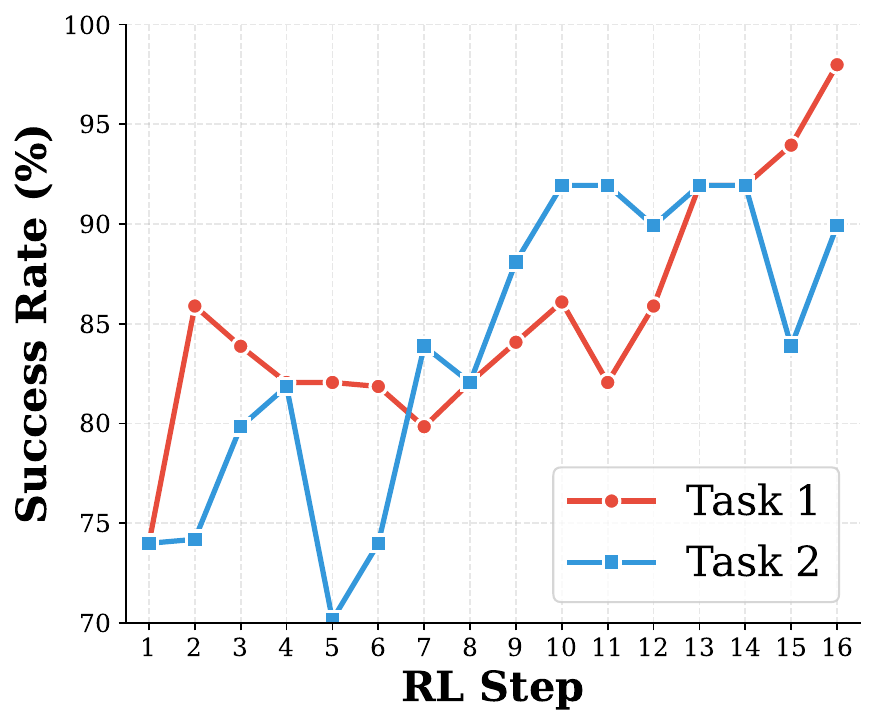}
        \caption{LIBERO-Object}
        \label{fig:libero_object}
    \end{subfigure}
    \hfill
    \begin{subfigure}[b]{0.24\textwidth}
        \centering
        \includegraphics[width=\linewidth]{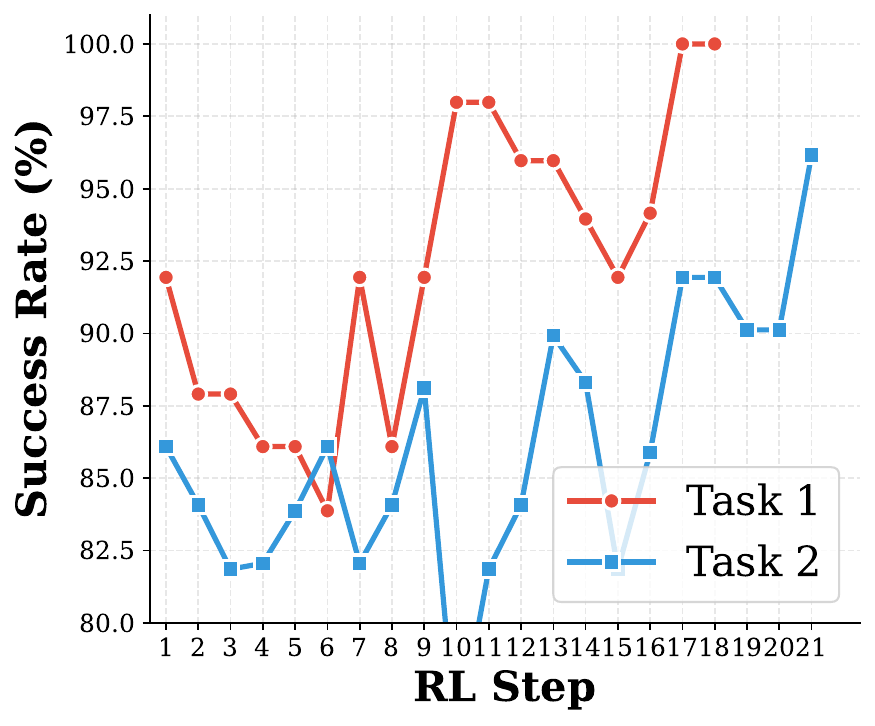}
        \caption{LIBERO-Goal}
        \label{fig:libero_goal}
    \end{subfigure}
    \hfill
    \begin{subfigure}[b]{0.24\textwidth}
        \centering
        \includegraphics[width=\linewidth]{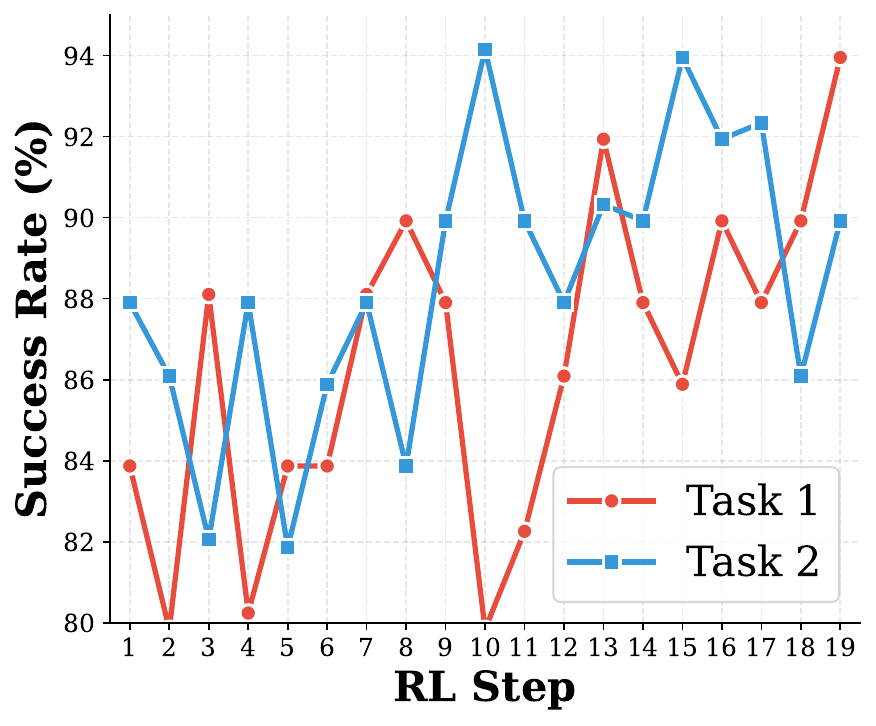}
        \caption{LIBERO-Spatial}
        \label{fig:libero_spatial}
    \end{subfigure}
    \hfill
    \begin{subfigure}[b]{0.24\textwidth}
        \centering
        \includegraphics[width=\linewidth]{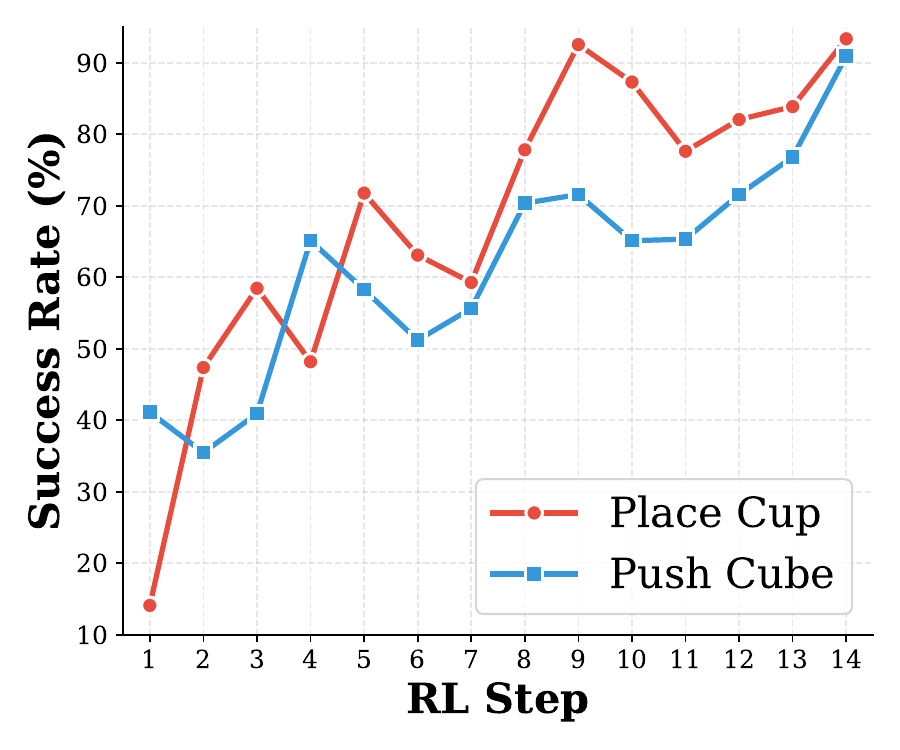}
        \caption{Real-World}
        \label{fig:real_world}
    \end{subfigure}
    
    \caption{Success rate improvements along \ourmethod RL training steps.}
    \label{fig:RL-curve}
    \vspace{-3mm}
\end{figure*}

We conduct experiments across three task suites from the LIBERO benchmark and tasks in real-world settings. Since LIBERO-100 primarily involves long-horizon tasks requiring the generation of over 300 video frames, where current autoregressive video models often suffer from severe quality drift, we leave its exploration for future work. For each task, 80–100 success and near-success trajectories are utilized to fine-tune the world model; among these, a subset of approximately 50 success trajectories is used to fine-tune the OpenVLA policy. As illustrated in Table~\ref{tab:vla-rl-improvements} and Figure~\ref{fig:RL-curve}, success rates across both the LIBERO suites and the real-world tasks exhibit significant improvements throughout the RL post-training process. To clarify, the real-world success rate curves in Figure~\ref{fig:RL-curve} are evaluated within the world simulator, whereas the final results reported in Table~\ref{tab:vla-rl-improvements} are obtained through physical, real-world experiments.

\textbf{Iterative Refinement through Policy Rollouts} As discussed in Section~\ref{method:world-simulator-deployment}, once a functional policy is established, its rollout trajectories can be leveraged to augment the \ourdataset dataset, which is then utilized to further refine the world model, enhancing its precision and action-following capabilities. Ultimately, this iterative cycle improves the overall efficiency of VLA reinforcement learning within the neural simulator. 

We conduct iterative experiments within our real-world setting. In the initial phase (Step 0), the SUPA dataset comprises manually collected success and near-success trajectories, alongside rollouts from the SFT OpenVLA-OFT baseline. For the next iteration (Step 1), the dataset is expanded to include newly generated trajectories from the RL-optimized policy produced in the previous step. Both world models are initialized from the ManiSkill-pretrained checkpoint, while the VLA policies for each RL stage begin from the base SFT version. As illustrated in Figure~\ref{fig:teaser} (b), this iterative refinement successfully boosts performance: the base SFT policy achieved only the success rate of 13.3\% and 26.7\%, after two RL iterations within our loop, it eventually reached 50.0\% and 53.3\% on two real-world tasks. These results demonstrate the overall effectiveness of our proposed closed-loop framework in enabling continuous policy and world model improvement.

\begin{wraptable}{r}{0.55\textwidth}
\vspace{-10pt} 
\centering
\caption{Ablation results. We investigate how different design choices influence the world model prediction accuracy.}
\label{tab:ablation}
\resizebox{0.54\columnwidth}{!}{%
\begin{tabular}{@{}lccc@{}}
\toprule
\textbf{Metric} 
& \multicolumn{2}{c}{\textbf{LIBERO-Object}} &
\multirow{2}{*}{\textbf{Real-World}} \\ 
\cmidrule(lr){2-3}
& \textbf{Task 1} & \textbf{Task 2} &  \\ 
\midrule

\multicolumn{4}{l}{\textbf{Visual Alignment}} \\
\makecell[l]{w/o near-success data} & 60\% & 66\% & 54\% \\
\makecell[l]{w/o reward prediction head} & 68\% & 70\% & 80\%  \\
ours & \textbf{92\%} & \textbf{90\%} & 90\% \\

\midrule
\multicolumn{4}{l}{\textbf{Reward Alignment}} \\
\makecell[l]{Qwen3-VL-8B-Instruct} & 84\% & 58\% & 84\% \\
ours & \textbf{88\%} & \textbf{90\%}  & \textbf{94\%}\\

\bottomrule
\end{tabular}
}
\vspace{-10pt} 
\end{wraptable}

\begin{figure*}[t]
  \centering
   \includegraphics[width=1\linewidth]{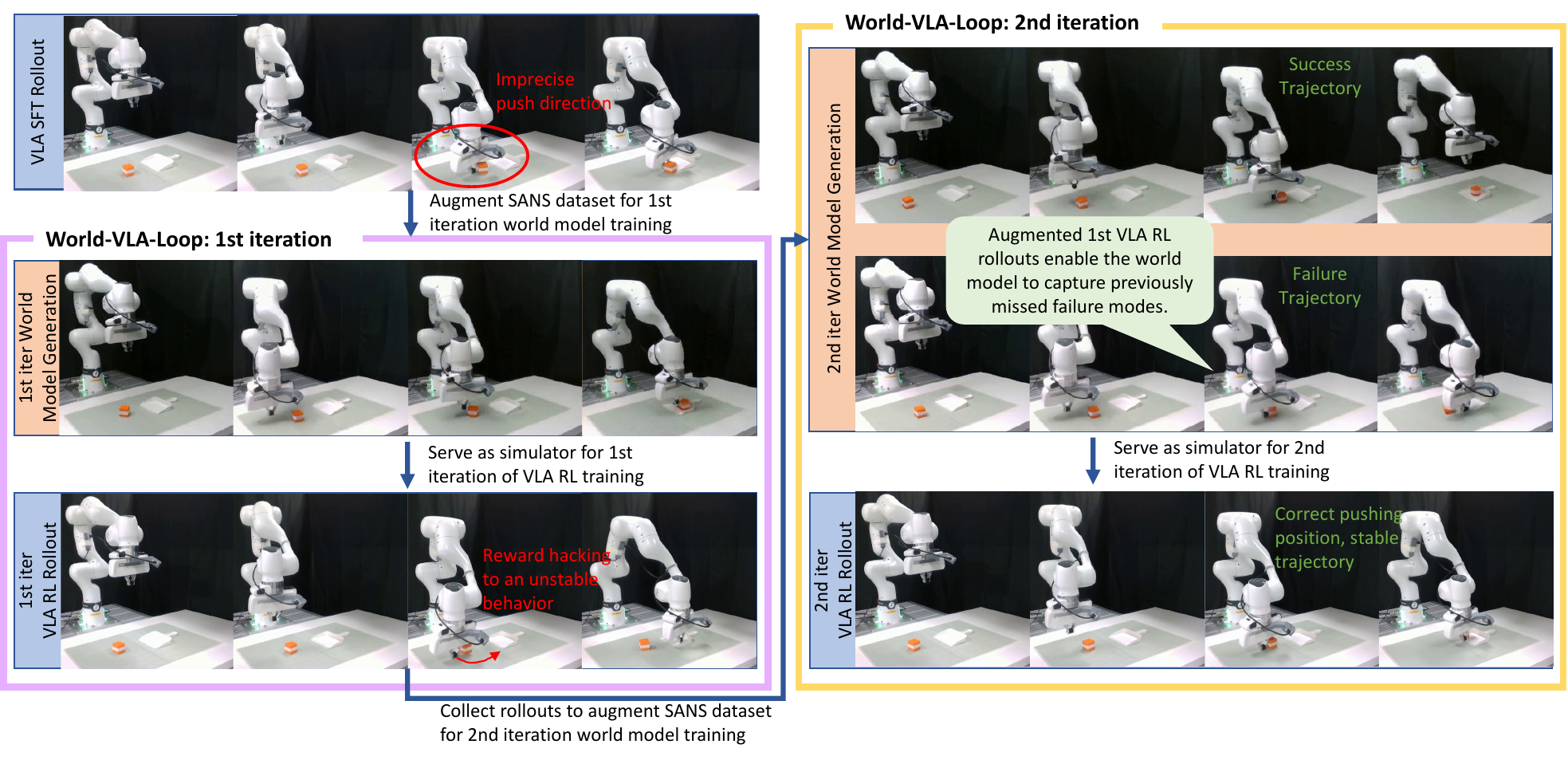}
    \vspace{-1mm}
   \caption{Examples of world model generated rollouts and actual execution videos by both SFT policy and RL post-trained policy. The policy rollouts augment the SANS dataset for world model training, which in turn serves as better simulators for RL iterations, leading to improved policies.}
   \label{fig:examples}
   \vspace{-3mm}
\end{figure*}

\subsection{Ablation Study}

\label{sec:ablation-study}
\textbf{Reward Prediction Head} We conduct ablation experiments to validate the effectiveness of the reward prediction head. First, we examine the world model's performance after removing this head. As indicated in Table~\ref{tab:ablation}, visual alignment drops by approximately 30\% without reward supervision, suggesting that joint training of the reward head and the diffusion backbone significantly improves the model's ability to discriminate between different action outcomes. Second, we compare our reward head against Qwen3-VL \cite{Qwen3-VL} acting as a success-failure judge—a common approach in prior literature. Specifically, we provide Qwen3-VL with the video frames generated by our world model and prompt it to evaluate the task's final success status. Notably, these are the same frames used to produce our internal reward predictions. The results in Table~\ref{tab:ablation} show that the VLM suffers from hallucination in specific domains and gets weaker accuracy, making it a less reliable reward function for RL compared to our method. While task-specific fine-tuning might improve VLM accuracy, it would significantly increase pipeline complexity. In contrast, our reward prediction head, which shares the backbone with the video generator, is not only more accurate but also enhances video quality through the benefits of joint training. For more results please refer to Appendix~\ref{sec:ablation-more}.

\textbf{Near-Success Data during World Model Training}
We also fine-tune the world model while excluding near-success trajectories. As indicated in Table~\ref{tab:ablation}, the visual alignment of the generated videos decreases significantly under this configuration. This confirms that near-success trajectories are indispensable for world model training, as they enable the model to internalize potential failure modes possible during VLA execution and more faithfully reflect resulting action outcomes.

\subsection{Qualitative Results}

Figure~\ref{fig:examples} presents qualitative examples of synthesized world model rollouts alongside actual VLA policy executions. Through our iterative refinement loop, the world model progressively learns to cover a broader action space and better represent the stochasticity of VLA exploration. In the policy rollouts, we observe that the SFT-baseline tends to generate imprecise pushing poses. While the first RL iteration mitigates this issue, the policy initially engages in reward hacking, converging on suboptimal behavior such as moving in front of the cube, since the first iteration world model fails to accurately model the spatial nuances of such trajectory. After augmenting the \ourdataset dataset, the world model develops a finer awareness of these edge cases. Consequently, the VLA policy trained within this refined environment achieves significantly more precise and robust grasping poses. For additional qualitative examples, please refer to Appendix~\ref{sec:supp-qualitative-examples}.

%% file: sec/5_conclusion.tex
\section{Conclusion}

\label{sec:conclusion}

We presented \ourmethod, a novel framework for world-model-based VLA reinforcement learning through the joint optimization of our world model and VLA policy. By leveraging the \ourdataset dataset curated with near-success trajectories, we trained a high-fidelity world model with integrated reward supervision, enabling effective policy refinement within a virtual environment. Furthermore, our co-evolving paradigm uses real-world rollouts to augment training data, progressively enhancing both world model grounding and policy performance. Experiments demonstrate significant success rate gains across simulation and real-world benchmarks, offering a scalable direction for real-world RL by reducing reliance on costly physical interactions.

\textbf{Limitations and Future Work}: Current autoregressive video models suffer from limited context memory and quality drift, which diminishes performance in long-horizon tasks exceeding 300 frames. Future research will explore integrating video backbones with enhanced long-term stability. Additionally, transitioning from sparse final-state rewards to step-wise intermediate sub-goals could further improve reward head accuracy and RL convergence.

%% file: sec/X_suppl.tex
\clearpage
\setcounter{page}{1}

\section{Implementation Details}
\label{sec:implementation-details}

\subsection{World Model Training}

We transfer the world model from Cosmos Predict 2 action-conditioned version checkpoint. As stated in the main paper, we first train the world model on a collected Maniskill \ourdataset dataset. This stage enables the model to grasp basic relationship between Franka robot arm and conditioning action, as well as some fundamental physics dynamics in manipulation scenarios. After this pretraining phase, we can adapt our model towards any downstream task with few data (less than 100 success and near-success trajectories). For the downstream task fine-tuning, we adopt a smaller learning rate and keep using full parameter tuning. For Maniskill pretraining we use the learning rate of $1e-3$, and for downstream tasks we use $1e-5$. We use FusedAdam as optimizer.

\subsection{VLA Deployment in World Model}

We adapt the SimpleVLA-RL codebase by replacing the physical simulator with our learned world model, while maintaining a consistent interface across both environments. The group size for GPRO to compute the advantage is set to 8. To facilitate the batch optimization required for GRPO, we implement the interaction between the VLA policy and the world model via a request-response architecture. The world model operates as a backend server that monitors incoming requests; once the VLA policy generates an action chunk, it is transmitted to the server, which allocates the task to an available model worker to compute the subsequent observation and reward signals. We use the exact same set of hyperparameters as SimpleVLA-RL \cite{li2025simplevla} except for the chunk size using 24.

\subsection{Real-World Experimental Settings}

\begin{figure}[h]
  \centering
   \includegraphics[width=0.9\linewidth]{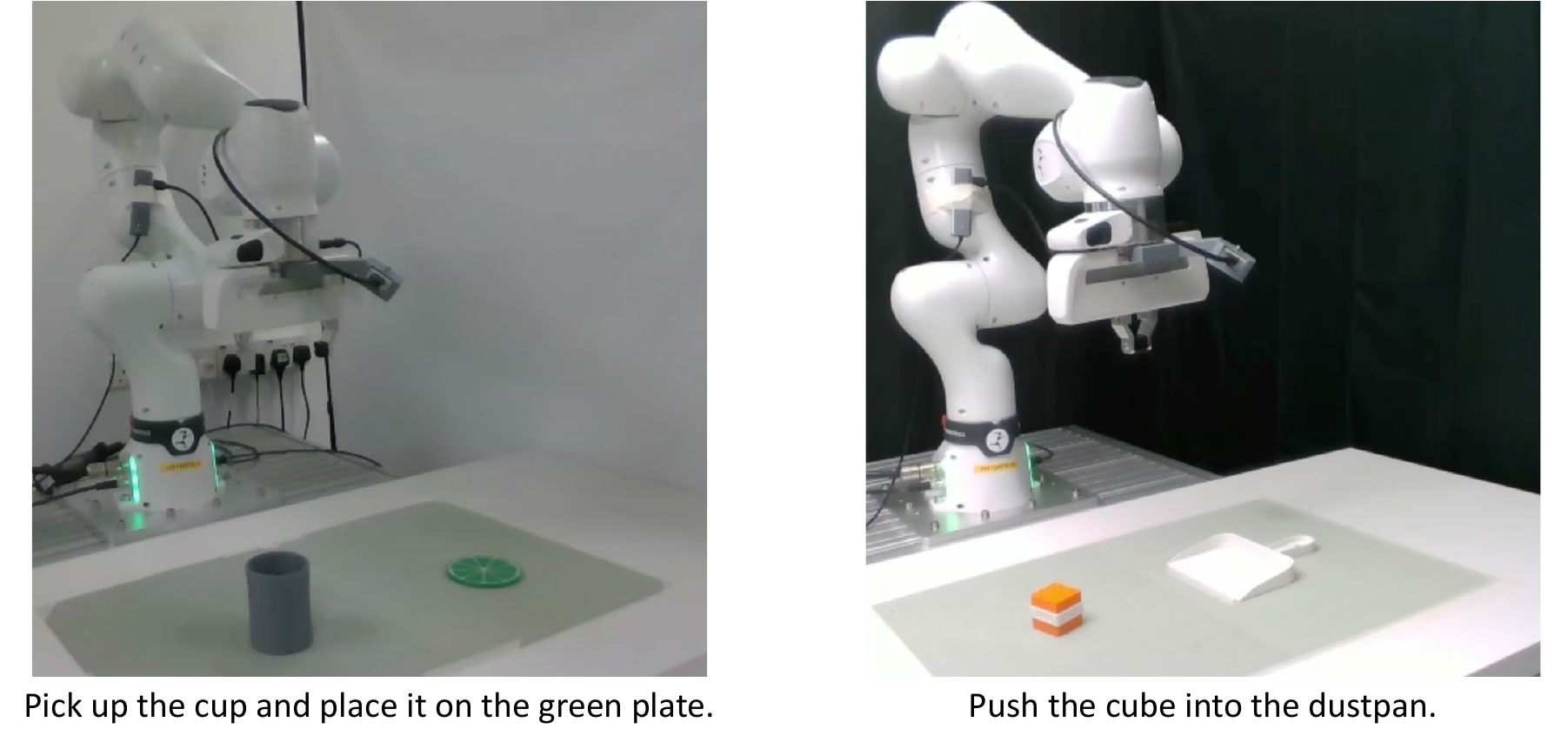}

   \caption{Real world experiments setup.}
   \label{fig:realworld-setup}
\end{figure}

We conducted two real-world experiments, covering common robot manipulation actions `pick and place' and `push'. As depicted in Figure ~\ref{fig:realworld-setup}, in our real-world experiments, the robot was tasked with the prompt \textit{"Pick up the cup and place it on the green plate."} and \textit{"Push the cube into the dustpan."} respectively. The positions of the cup, cube and plate were randomized. These tasks are challenging for VLA policies as it requires rigorous grasping and pushing precision, otherwise in the first task, the rounded cup easily slips away, and in the second task the gripper tends to push crookedly.

\subsection{Computation Cost}

\textbf{Inference Efficiency} A batched world model generation of 24 video frames takes approximately 6 seconds on a single NVIDIA H100 node. In our experiments, a full RL training session for a task typically converges within 20–40 optimization steps, totaling roughly 25 to 35 hours. This represents a significant improvement over traditional physical rollouts, which require labor-intensive manual resets and constant human supervision.

\textbf{Training Efficiency} World model fine-tuning on a specific task with ~100 trajectories typically converges within 4 hours on 8 H100 GPUs, making the approach highly affordable and scalable for new task adaptation.

\section{Additional Ablation}
\label{sec:ablation-more}

\subsection{Quality Analysis on Long-Horizon Tasks}

To clarify on the long-horizon tasks, we measure the video quality of world model in different frame length:

\begin{table}[h]
\centering
\caption{Performance Metrics across Frames}
\label{tab:frame_metrics}
\begin{tabular}{ccc}
\hline
\textbf{Frame} & \textbf{PSNR} & \textbf{SSIM} \\
\midrule
200            & 23.5890       & 0.7430        \\
250            & 22.2148       & 0.6731        \\
300            & 20.2351       & 0.6282        \\
\bottomrule
\end{tabular}
\end{table}

Quality analysis confirms that severe degradation typically occurs only after the 300-frame mark, which is sufficient for the majority of robotic manipulation tasks. LIBERO-Object, Spatial and Goal tasksuites typically require around 220 frames (10 passes), LIBERO-Long tasks extend to 300-400 frames (15 passes) and thus the video becomes too blurry for RL accuracy in the task success section of these tasks. In the future, we will leverage newer auto-regressive video generation techniques and we believe this limitation will be finally addressed.

\subsection{VLM Reward for RL}

In Section ~\ref{tab:ablation}, we compared the world model generation accuracy of using VLM and reward prediction head. Here we also conducted RL experiments on LIBERO-Object using a VLM-based reward (Qwen3.5-VL-Instruct 8B). We replaced our integrated reward head with the VLM, feeding generated video frames to the VLM with prompts to output binary success states. The results are shown below:

\begin{table}[h]
\centering
\caption{Performance comparison on object tasks.}
\label{tab:object_results}
\begin{tabular}{@{}lcc@{}}
\toprule
\textbf{Model} & \textbf{Object-1} & \textbf{Object-2} \\
\midrule
Ours & 97.9\% & 91.9\% \\
Ours w/ VLM & 93.9\% & 88.0\% \\
\bottomrule
\end{tabular}
\end{table}

This experiment demonstrates two key findings: (1) High-quality video generation: The fact that VLM rewards achieve reasonable performance validates that our world model generates sufficiently realistic and accurate videos that can be effectively interpreted by external VLMs. (2) Efficiency advantage: Our integrated reward head achieves superior performance while being significantly more computationally efficient than VLM-based alternatives. This confirms that joint reward supervision successfully enables the world model to distinguish between success and failure rollouts more accurately than external reward functions.


\section{Visual Examples}
\label{sec:supp-qualitative-examples}

\subsection{Additional World Model Generation Results}

\begin{figure*}[h]
  \centering
   \includegraphics[width=0.95\linewidth]{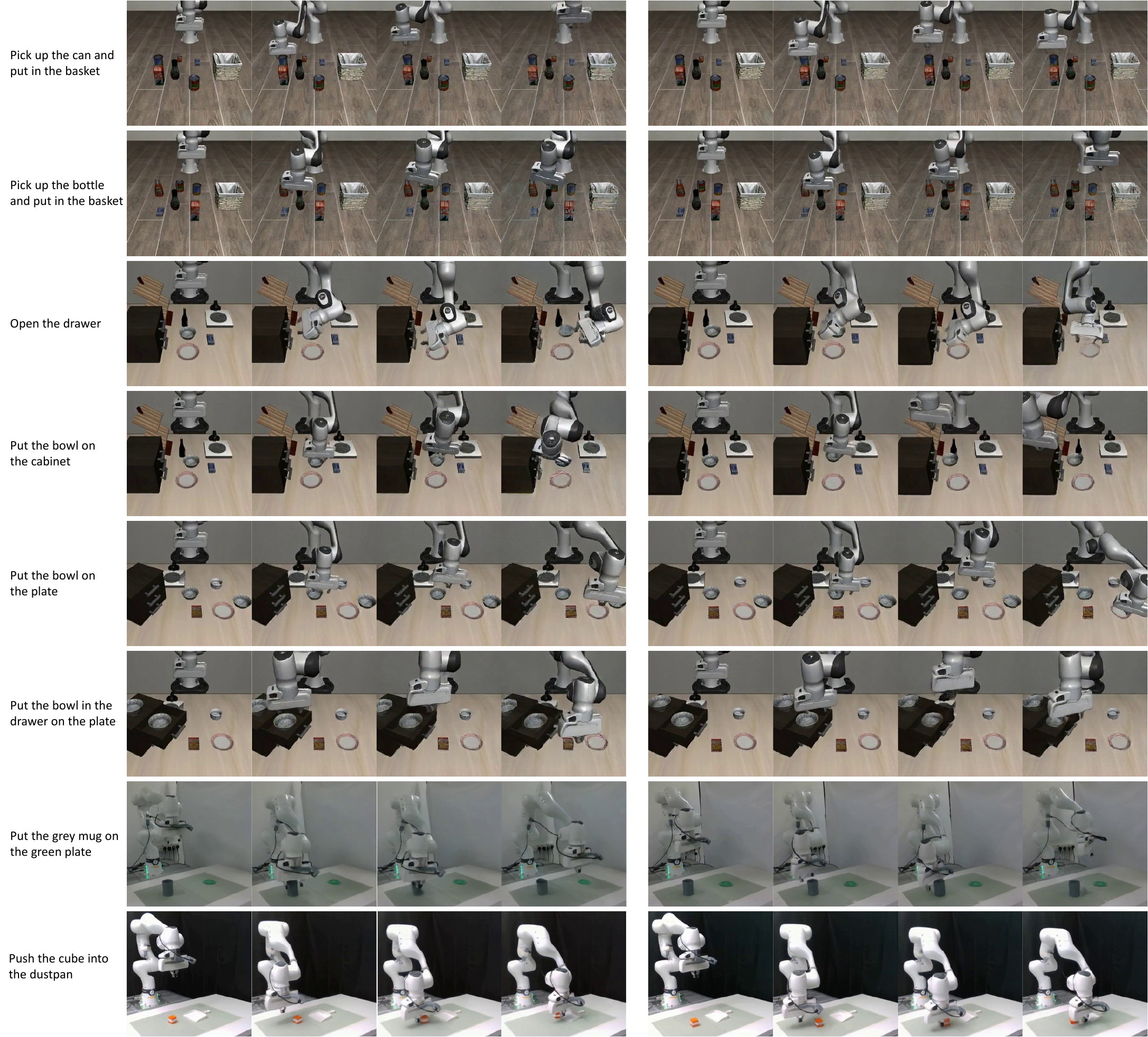}

   \caption{Examples of world model generated rollouts of failure trajectories.}
   \label{fig:examples-fail-supp}
\end{figure*}

\begin{figure*}[h]
  \centering
   \includegraphics[width=0.95\linewidth]{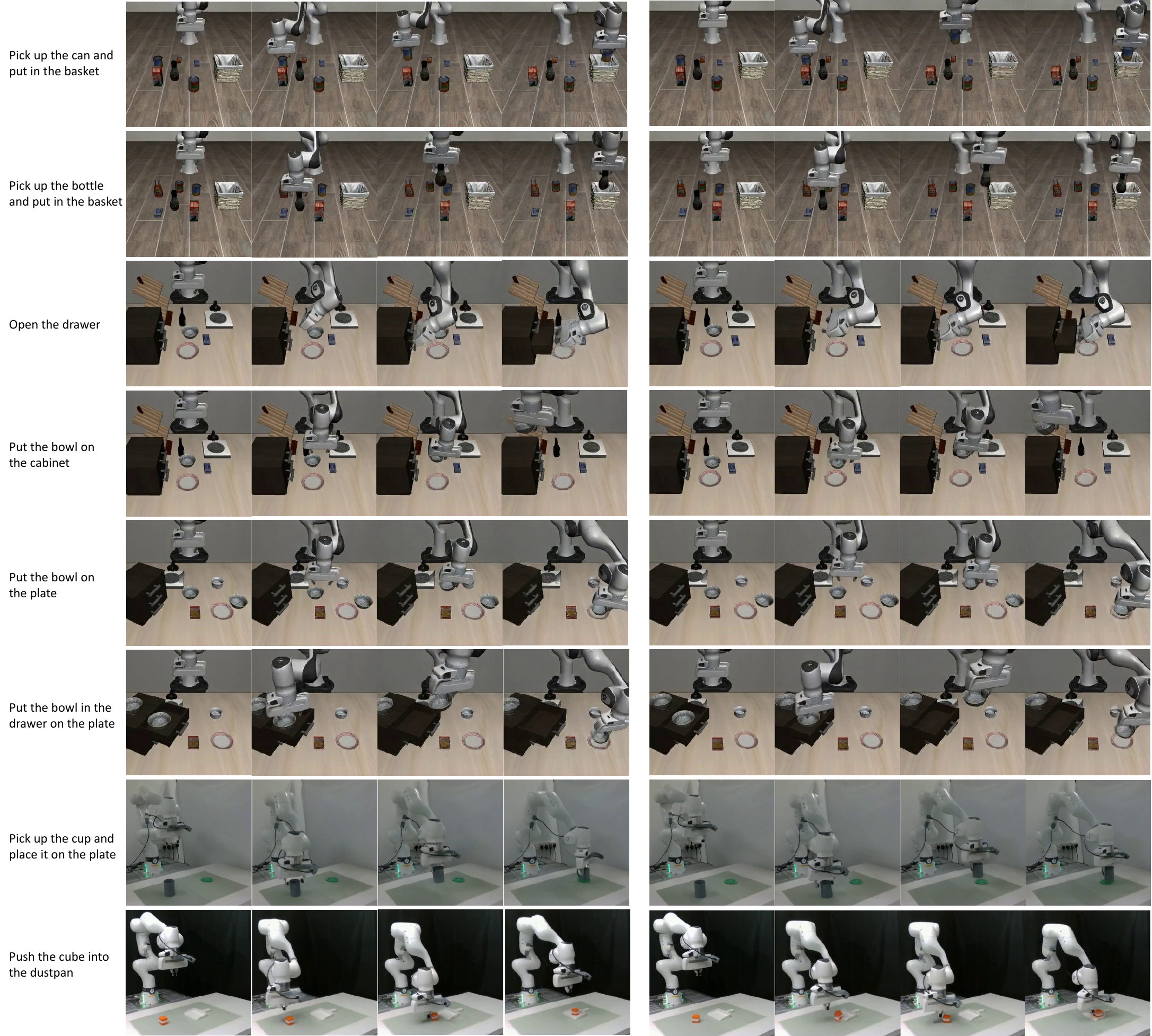}
   \caption{Examples of world model generated rollouts of successful trajectories.}
   \label{fig:examples-succ-supp}
\end{figure*}

We provide additional qualitative examples of our world model's generation capabilities in this section. Figure~\ref{fig:examples-fail-supp} illustrates failure rollouts synthesized by the world model when conditioned on erroneous action trajectories. The generated video frames demonstrate that the world model accurately adheres to the conditioning actions and precisely distinguishes between nuanced outcomes resulting from incorrect execution. This capability ensures that the model covers a diverse range of failure modes, providing the necessary grounding for effective policy refinement. Figure~\ref{fig:examples-succ-supp} provides some examples of generated successful trajectories. Most of the videos show that world model can clearly distinguish between failure and success action inputs.

\subsection{World Model in Unseen Cases}

\begin{figure*}[t]
  \centering
   \includegraphics[width=0.95\linewidth]{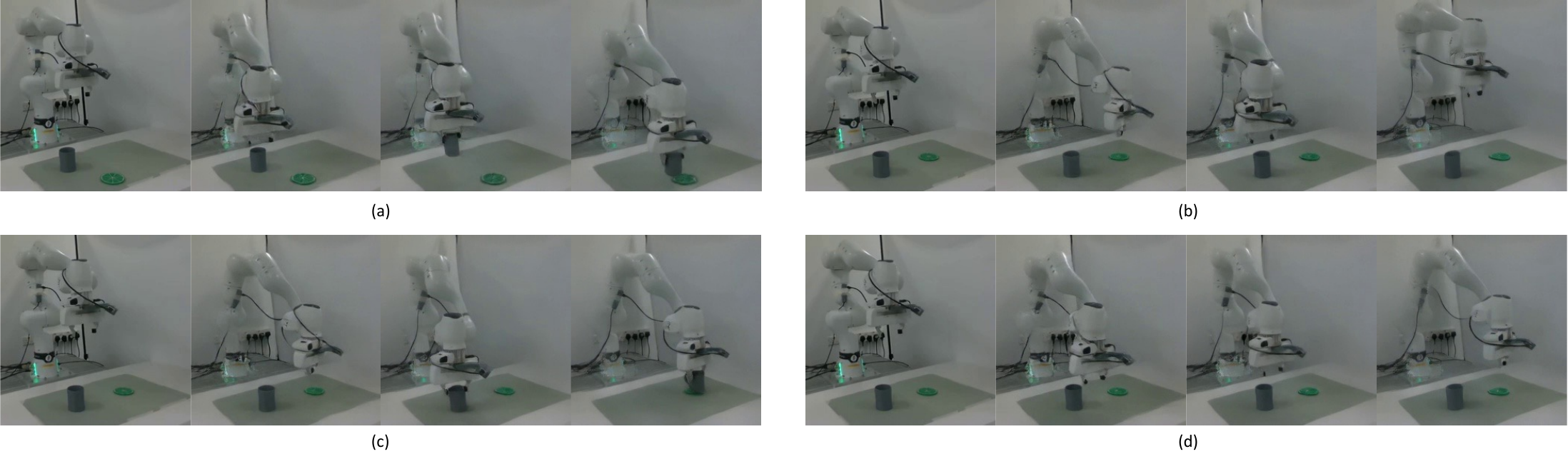}
   \caption{World model generation results on unseen action sequences. (a) The plate is initialized directly in front of the mug. (b) The gripper moves to the right, returns to the mug's overhead position, and retracts to a neutral pose. (c) The gripper first moves to the right, followed by a successful pick-and-place of the mug onto the plate. (d) A sequence of forward, backward, and lateral oscillations.}
   \label{fig:ood_wm}
\end{figure*}

Notably, our world model demonstrates the ability to synthesize trajectories unseen during the fine-tuning of the target \ourdataset dataset, leveraging the broad priors learned during its pretraining on ManiSkill. Figure~\ref{fig:ood_wm} illustrates cases where the world model is conditioned on entirely novel action sequences and robot arm trajectories. Despite the stochastic nature of these unseen inputs, the model faithfully adheres to the action conditioning, generating physically plausible motion aligning with ground truth. This robustness indicates that the world model has effectively internalized the underlying mapping between control actions and robotic kinematics, rather than merely memorizing training sequences.

